\PassOptionsToPackage{table}{xcolor}
\documentclass[letterpaper]{article} 
\usepackage[preprint]{aaai2027}  
\usepackage[hyphens]{url}  
\usepackage{graphicx} 
\usepackage{natbib}  
\usepackage{caption} 
\usepackage{algorithm}
\usepackage{algorithmic}

\usepackage{newfloat}
\usepackage{listings}
\DeclareCaptionStyle{ruled}{labelfont=normalfont,labelsep=colon,strut=off} 
\floatstyle{ruled}
\newfloat{listing}{tb}{lst}{}
\floatname{listing}{Listing}

\usepackage{booktabs}
\usepackage{multirow}
\usepackage{amssymb}
\usepackage{amsmath}
\usepackage{hyperref}

\title{Private Face Recognition Training Dataset Publication via Identity-Decoupled and Geometry-Preserving Face Distillation}
\author{
    Shuhuan Chen\textsuperscript{\rm 1,3}, Xiangyu Zhu\corresponding\textsuperscript{\rm 2,4}, Weisong Zhao\textsuperscript{\rm 5}, Siran Peng\textsuperscript{\rm 2,4}, Tianshuo Zhang\textsuperscript{\rm 2,4},\\Haoyuan Zhang\textsuperscript{\rm 2,4}, Haichao Shi\textsuperscript{\rm 1}, Xiao-Yu Zhang\corresponding\textsuperscript{\rm 1}, Zhen Lei\textsuperscript{\rm 2,4,6,7}\\
}
\affiliations{
    \textsuperscript{\rm 1}IIE, CAS
    \textsuperscript{\rm 2}CBSR\&MAIS, CASIA
    \textsuperscript{\rm 3}SCS, UCAS
    \textsuperscript{\rm 4}SAI, UCAS
    \textsuperscript{\rm 5}Sangfor
    \textsuperscript{\rm 6}CAIR, HKISI, CAS
    \textsuperscript{\rm 7}SCSR, FIE, M.U.S.T\\
    \{chenshuhuan, zhangxiaoyu\}@iie.ac.cn, \{xiangyu.zhu, zhen.lei\}@ia.ac.cn
}

\begin{document}

\maketitle

\begin{abstract}
Publishing private face recognition~(FR) training datasets is privacy-sensitive because faces expose identity information. Private FR training dataset publication mitigates this risk by releasing protected proxies as substitutes for private training faces. However, training FR models with such data introduces an identity paradox: \emph{the identity cues that make released faces useful for recognition supervision are also the cues that make them linkable to real individuals.} A protected face should be decoupled from the original identity, yet still behave as a reliable identity sample for training. Removing these cues too aggressively may destroy the class structure needed for recognition learning, whereas preserving them too faithfully may increase source-identity linkability. We argue that this paradox stems from conflating source-aligned identity semantics with recognition-useful proxy identity geometry. The former should be suppressed to reduce linkage to private individuals, while the latter should be preserved for FR learning. Based on this insight, we propose \textbf{Private Face Distillation}, an identity-decoupling and geometry-preserving framework. It uses Orthogonal Geometry Preservation to construct decoupled proxy identities from private identity representations while maintaining hyperspherical geometry, and Relational Topology Alignment to preserve identity relations for recognition learning. Experiments across multiple domain-shifted FR scenarios show that Private Face Distillation achieves stronger utility than the evaluated publication baselines. On IJB-C surveillance, it improves $\mathrm{TAR}@\mathrm{FAR}{=}1\text{e-}{3}$ by 3.94\% over the baseline while reducing source-identity linkability. These results suggest that private FR training dataset publication should decouple source-identity correspondence while preserving proxy identity geometry.~\href{https://github.com/MrHuan3/PrivateFaceDistillation}{Codes.}
\end{abstract}


\section{Introduction}
\label{sec:intro}
High-quality face recognition training datasets are crucial for improving model performance in domain-specific scenarios, yet their publication is increasingly restricted by identity privacy concerns~\cite{10.1145/3708501}. Unlike generic visual data, facial images directly encode biometric identity, making released datasets vulnerable to unauthorized re-identification, identity tracing, and even deepfake misuse~\cite{9481149,NEURIPS2025_062cbe9b,Peng_2026_CVPR}. Once publicly distributed, private faces may be persistently linked across datasets, applications, or surveillance scenarios, creating risks that are difficult to revoke. To address this issue, Face Dataset Publication~\cite{11146914} has recently been proposed as a privacy-aware data release paradigm, aiming to transform a private face dataset into a protected counterpart for downstream model training. This work studies its FR-specific setting, where identity cues are required for recognition supervision yet also enable linkage between released faces and source individuals.
\begin{figure}[t]
	\centering
	\includegraphics[width=0.95\columnwidth]{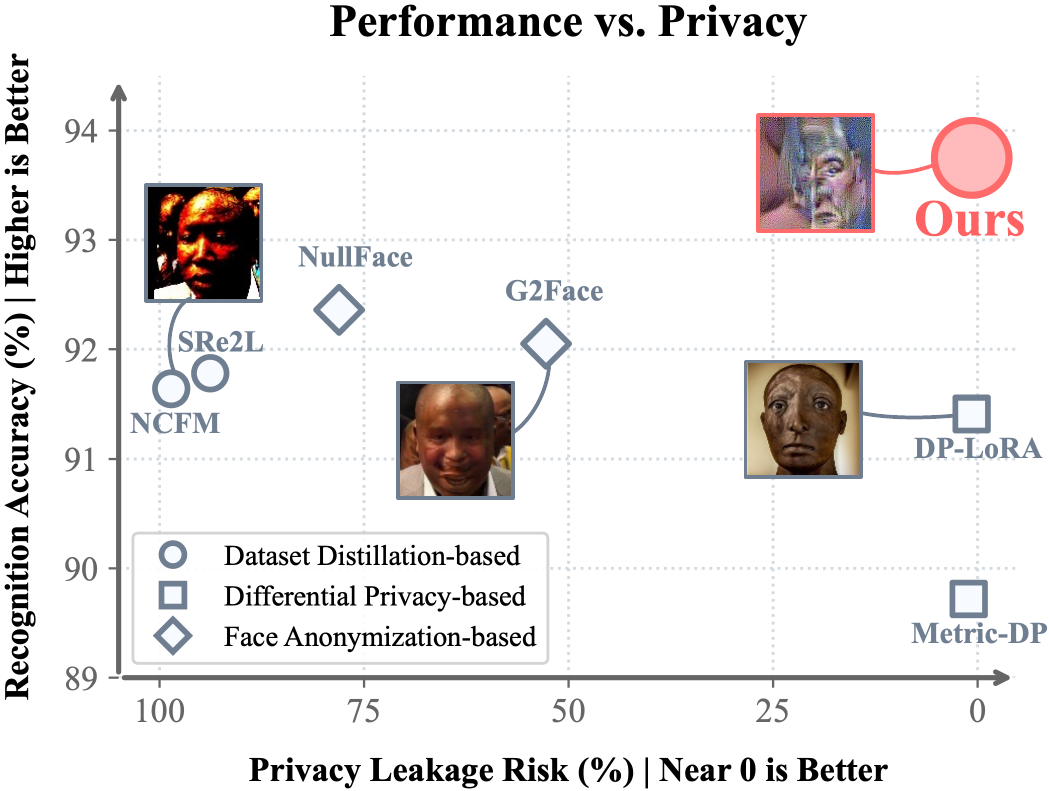}
	\caption{Performance-linkability trade-off on IJB-C surveillance under private FR training dataset publication. PFD improves downstream utility while reducing source-identity linkability under the evaluated attack protocol.}
	\label{fig:teaser}
\end{figure}
Existing methods typically weaken the association between released faces and private individuals through face anonymization or differential privacy. Face anonymization~(FA) modifies or synthesizes facial appearances to reduce their visual correspondence to the original individuals~\cite{10.1145/3474085.3475464,10.1145/3503161.3548202,10.1145/3474085.3475367}. While this can reduce apparent identity similarity, it mainly changes who the face looks like, without specifying how protected samples should behave as identity classes in the recognition space. Differential privacy~(DP) provides theoretically grounded protection by limiting the influence of individual samples or identities through perturbation or noise injection~\cite{Zhu_2020_CVPR,meng2022improving}. However, perturbations that support privacy guarantees also modify the training signals in released data, potentially obscuring subtle identity differences needed for recognition learning. Moreover, Face Dataset Publication often relies on limited private data, making domain-specific generative proxy synthesis difficult. These limitations reveal a fundamental identity paradox when published data are used for FR training: \emph{the same identity cues that make protected faces useful for recognition supervision also make them linkable to real individuals.} A protected face should reduce correspondence to its source identity while remaining a reliable identity sample for recognition training. Suppressing identity cues too strongly may weaken class consistency and separability, while preserving them too faithfully may increase source-identity linkability.

We argue that this paradox reflects not an inherent incompatibility between privacy protection and recognition learning, but a conflation of two forms of identity information. Source-aligned appearance and embedding cues link released faces to real individuals and increase privacy risk. Recognition learning, by contrast, depends on class geometry, including intra-class consistency, inter-class separability, and identity relations, without preserving the original identity correspondence. Private FR training dataset publication should therefore weaken source-identity linkability while preserving proxy identity geometry. Released faces need not reproduce the original individuals, but should retain how proxy classes are organized for recognition learning.

Motivated by this objective, we transform private identity representations into protected proxy identities while preserving recognition-useful geometry. We instantiate \textbf{Private Face Distillation}~(\textbf{PFD}), an identity-decoupling and geometry-preserving framework for private FR training dataset publication. PFD separates source-aligned identity correspondence from recognition-useful proxy geometry through two designs. Orthogonal Geometry Preservation constructs a transformed space with distinct proxy class targets while preserving hyperspherical identity relations. Relational Topology Alignment transfers this geometry to releasable images by aligning intra-class and inter-class relations, enabling stable, discriminative proxy classes for training. Only proxy images and arbitrary class labels are released, while source data, the orthogonal transformation, and adapted model remain local to the publisher. As shown in Fig.~\ref{fig:teaser}, PFD achieves a favorable performance-linkability trade-off, improving downstream FR utility while reducing source-identity linkability under the evaluated attacks. Our contributions are threefold:

\begin{itemize}
	\item We identify the identity paradox in private FR training dataset publication, where source-aligned correspondence increases linkability while proxy identity geometry supports recognition learning.
	\item We introduce Private Face Distillation, which combines OGP and RTA to construct identity-decoupled RGB proxies while preserving recognition-useful geometry.
	\item Across four domain shifts, PFD achieves the highest average utility among the evaluated publication methods while maintaining low source-identity linkability under the tested attacks.
\end{itemize}

\section{Related Work}
\label{sec:relat}

\subsection{Face Dataset Publication}
Face Dataset Publication aims to transform a private face dataset into a protected counterpart that can be released for downstream training while reducing identity privacy risks~\cite{11146914}. Existing methods mainly rely on differential privacy or face anonymization. Differential-privacy-based approaches~\cite{Chen_2022_CVPR,Tsai_2025_ICCV} provide theoretically grounded protection through perturbation or privacy-constrained optimization, but do not explicitly specify what identity supervision should remain for FR adaptation. Face anonymization methods~\cite{Barattin_2023_CVPR,Kuang_2024_CVPR,10499238} reduce apparent identity similarity by synthesizing facial appearances, but visual identity replacement does not directly define how protected samples should be organized in the recognition space. In contrast, our work focuses on its FR training-data setting, aiming to reduce source-identity correspondence while preserving proxy identity geometry.

\subsection{Dataset Distillation}
Dataset distillation~(DD) aims to synthesize a compact proxy set that can substitute for a large real dataset in downstream training~\citep{wang2018dataset}. Existing methods typically construct such proxies through optimization-based or generation-based routes~\cite{Cazenavette_2022_CVPR,chen2025influenceguided,NEURIPS2025_615ce9f0,chan2025mgd3,11509640}. This formulation is naturally relevant to Face Dataset Publication, as both settings rely on proxy data to replace real training data. However, conventional DD focuses on training utility and does not require the proxies to protect private identities. Prior work shows that distilled proxies may provide implicit privacy effects~\cite{pmlr-v162-dong22c}, suggesting their potential for privacy-aware data release. Yet such privacy is only a by-product and does not address the identity paradox, where identity information is both the supervision signal and the privacy risk. Our work adapts DD's proxy-data advantage to private FR training dataset publication by reducing source-identity correspondence while preserving recognition-useful proxy identity geometry.

\subsection{Distinction from PPFR}
Privacy-preserving face recognition~(PPFR) generally protects facial inputs or representations used during recognition~\cite{Wang_2023_CVPR,Mi_2023_ICCV,Mi_2024_CVPR,daifracface}. Private FR training dataset publication instead focuses on releasing protected data for reuse in downstream training. Such data should remain usable without access to the publisher's private components. PFD only releases labeled RGB proxy images, while the source data, learned transformation, and adapted publisher model remain private. A third party can train an FR model on these proxies using a standard pipeline, and the resulting model directly accepts ordinary RGB test images without protected-domain conversion or architectural modification. We therefore evaluate PFD as a published training resource rather than as a protected recognition system, considering both downstream recognition utility and source-identity linkability in the released proxy set.

\begin{figure*}[t]
	\centering
	\includegraphics[width=0.90\textwidth]{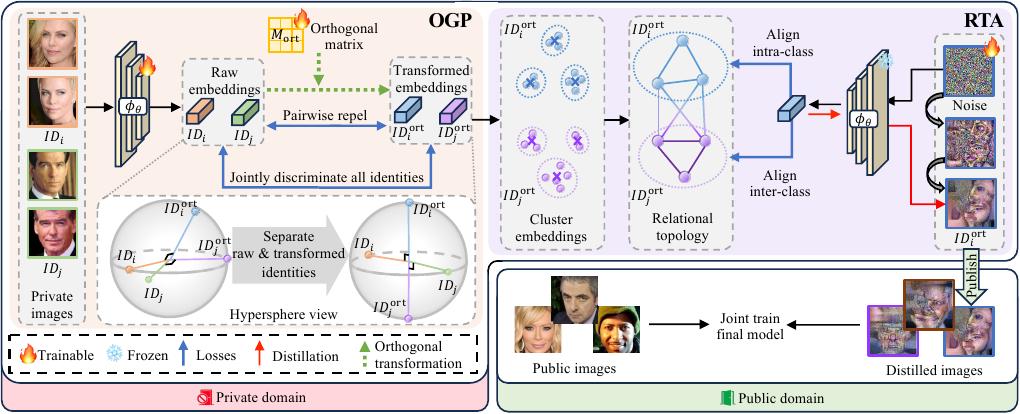}
	\caption{Overview of PFD. OGP learns an internal transformed identity space by orthogonally mapping raw embeddings, separating original and transformed identities, and preserving hyperspherical geometry. RTA distills proxy images from random noise by aligning the intra-class and inter-class topology of transformed embeddings. Only the distilled proxy images and arbitrary labels are published as a reusable FR training set for joint training with public data.}
	\label{fig:pl_mt}
\end{figure*}

\section{Method}
\label{sec:method}
\subsection{Data Release Setting}
In the private FR training dataset publication setting~\cite{11146914}, we release only protected RGB proxies and arbitrary class labels for downstream FR training. Private source images and internal states, including the source-to-proxy correspondence, adapted model, orthogonal matrix, and paired original and transformed embeddings, are not released. Once distillation and publisher-side evaluation are complete, these intermediate states are no longer needed and can be discarded without affecting the proxy set. The released proxy set requires no publisher-side components. The method and training protocol may be known, and the proxies may be analyzed using auxiliary identity images, public FR models, or off-the-shelf reconstruction models. We focus on whether released images can be linked to source identities while supporting FR training. Disclosure of retained publisher-side information is outside this setting.

\subsection{Private Face Distillation}
\label{sec:overview}
As illustrated in Fig.~\ref{fig:pl_mt}, PFD distills a protected RGB proxy training set by transferring recognition-useful identity geometry from private target-domain data into releasable proxy images. It first learns an internal transformed proxy space through \textbf{Orthogonal Geometry Preservation (OGP)}, which retains hyperspherical identity relations while reducing direct correspondence to the original source coordinates. Based on this space, \textbf{Relational Topology Alignment (RTA)} distills proxy images from random noise by aligning their representations with the intra-class and inter-class topology of transformed identity prototypes.

\subsection{Orthogonal Geometry Preservation}
\label{sec:ogp}
Directly distilling private embeddings may transfer source-aligned identity semantics into published proxies, while removing identity structure would destroy the angular geometry needed for FR adaptation. OGP resolves this by mapping private identity representations into a proxy space that preserves recognition-useful hyperspherical geometry while reducing direct correspondence to source identity coordinates.

\subsubsection{Geometry-Preserving Identity Decoupling.}
Margin-based FR methods~\cite{Sun_2020_CVPR,Boutros_2022_CVPR,10.1145/3581783.3611711,10914510} learn $\ell_2$-normalized embeddings on a hypersphere $\mathcal{M}\subset\mathbb{R}^D$, where identity discrimination mainly depends on angular relations. Given a private target-domain image $\hat{x}$, we compute its normalized embedding as $\hat{e}=\mathrm{norm}_{\ell_2}(\phi_{\theta}(\hat{x}))\in\mathcal{M}$, where $\phi_{\theta}$ is a trainable FR backbone initialized from a public pre-trained model and jointly optimized with OGP. To preserve this recognition-useful geometry without exposing the original identity coordinates, OGP maps each target-domain embedding $\hat e$ into a proxy space through an orthogonal transformation:
\begin{equation}
	\label{eq:lie}
	\hat e_{\mathrm{ort}}=M_{\mathrm{ort}}\hat e,\quad
	M_{\mathrm{ort}}=\exp(\Omega),\quad
	\Omega=A-A^\top,
\end{equation}
where $A\in\mathbb{R}^{D\times D}$ is learnable and $\Omega\in\mathfrak{so}(D)$ is skew-symmetric. The matrix exponential gives $M_{\mathrm{ort}}\in SO(D)$, allowing OGP to learn a strictly orthogonal transformation without heuristic orthogonality penalties or post-projection~\cite{pmlr-v97-lezcano-casado19a}. For a column-wise normalized prototype matrix $P\in\mathbb{R}^{D\times N}$ and $P_{\mathrm{ort}}=M_{\mathrm{ort}}P$, orthogonality preserves the within-space topology, $P_{\mathrm{ort}}^\top P_{\mathrm{ort}}=P^\top P$, whereas source-to-proxy correspondence is measured by $P^\top P_{\mathrm{ort}}=P^\top M_{\mathrm{ort}}P$. OGP therefore retains the hyperspherical geometry required for FR while defining transformed coordinate targets for proxy identities. Further analysis of this distinction and its privacy implications is provided in the Supplementary Materials.

\subsubsection{Discriminative Proxy Identity Learning.}
Although the orthogonal transformation preserves hyperspherical geometry, it should not collapse to the identity mapping or remain too close to the original identity coordinates. To construct non-trivial proxy identities, we train original and transformed identities jointly in the same FR embedding space. For $N$ private identities, we build a classifier head $h_{\theta}$ with $2N$ prototypes, where the first $N$ classes correspond to original identities and the remaining $N$ classes correspond to transformed proxy identities. Given an embedding $\hat e$ with label $\hat y$, its transformed counterpart $\hat e_{\mathrm{ort}}$ is assigned label $y = \hat y+N$. The classification loss is written as
\begin{equation}
	\mathcal{L}_{\mathrm{cls}}^{\mathrm{OGP}} = \mathcal{L}_{\mathrm{CE}} \Big(h_{\theta}\big([\hat e,\hat e_{\mathrm{ort}}]\big), [\hat y,y]\Big),
\end{equation}
where $[\cdot,\cdot]$ denotes batch-wise concatenation and $\mathcal{L}_{\mathrm{CE}}$ is a margin-based cross-entropy loss. This objective treats original and transformed identities as distinct classes while keeping them in a shared hyperspherical FR space. To further reduce direct correspondence to the original identity coordinates, we impose a repulsion constraint:
\begin{equation}
	\label{eq:rep}
	\mathcal{L}_{\mathrm{rep}}=\mathbb{E}_{\hat{e}\sim\mathcal{B}} \Big[\max\big(0, |\cos(\hat{e}_{\mathrm{ort}},\hat{e})|-m\big)\Big],
\end{equation}
where $\mathcal{B}$ is the current mini-batch and $m$ is a cosine margin. Meanwhile, to keep the proxy classifier consistent with the orthogonal geometry, we enforce prototype isomorphism:
\begin{equation}
	\label{eq:iso}
	\mathcal{L}_{\mathrm{iso}}=\frac{1}{N}\sum_{i=1}^{N}\left[1-\cos\left(P_{\mathrm{ort}}^{(i)},M_{\mathrm{ort}}P_{\mathrm{real}}^{(i)}\right)\right],
\end{equation}
where $P_{\mathrm{real}},P_{\mathrm{ort}}\in\mathbb{R}^{D\times N}$ collect the normalized original and transformed class prototypes of $h_{\theta}$ column-wise. The overall OGP objective is
\begin{equation}
	\label{eq:ogp}
	\mathcal{L}_{\mathrm{OGP}} = \mathcal{L}_{\mathrm{cls}}^{\mathrm{OGP}} + \lambda_{\mathrm{rep}}\mathcal{L}_{\mathrm{rep}} + \lambda_{\mathrm{iso}}\mathcal{L}_{\mathrm{iso}}.
\end{equation}
Through this objective, OGP separates source and proxy identities at the class level while preserving their within-space relational geometry. The complete stage therefore defines discriminative proxy coordinate targets rather than treating an orthogonal rotation alone as a privacy mechanism. The resulting backbone $\phi_{\theta}$ and classifier head $h_{\theta}$ provide the internal proxy embedding space for subsequent distillation, but are not released with the published dataset.

\subsection{Distillation via Relational Topology Alignment}
\label{sec:rta}
RTA converts the proxy identity geometry learned by OGP into releasable proxy images. Following the DD paradigm~\cite{NEURIPS2023_e91fb65c}, each proxy image $s$ is initialized from Gaussian noise and optimized with the adapted backbone $\phi_{\theta}$ and classifier head $h_{\theta}$. Rather than directly matching private embeddings or facial appearances, RTA distills relational topology to preserve intra-identity consistency and inter-identity relations. We further use Batch Normalization regularization to align proxy images with the feature statistics of $\phi_{\theta}$:
\begin{equation}
	\label{eq:bn}
	\mathcal{L}_{\mathrm{bn}} = \sum_{l} \left( \left \| \mu_l(s) - \mathbf{BN}_l^{\mathrm{RM}} \right \|_2 + \left \| \sigma_l^2(s) - \mathbf{BN}_l^{\mathrm{RV}} \right \|_2 \right),
\end{equation}
where $\mu_l(\cdot)$ and $\sigma_l^2(\cdot)$ are the batch mean and variance at layer $l$, and $\mathbf{BN}_l^{\mathrm{RM}}$ and $\mathbf{BN}_l^{\mathrm{RV}}$ are the running statistics of $\phi_{\theta}$. We also impose a classification loss:
\begin{equation}
	\label{eq:ce}
	\mathcal{L}_{\mathrm{cls}}^{\mathrm{RTA}} = \mathcal{L}_{\mathrm{CE}}\Big(h_{\theta}\big(\phi_{\theta}(s)\big), y\Big),
\end{equation}
where $y$ is the internal transformed-class label, consistently remapped to an arbitrary proxy label before release. The loss encourages discriminative proxy classes in the FR space.

\subsubsection{Relational Topology Alignment.}
Although $\mathcal{L}_{\mathrm{bn}}$ and $\mathcal{L}_{\mathrm{cls}}^{\mathrm{RTA}}$ make proxy images compatible with the adapted FR model $\phi_{\theta}$, they do not preserve the identity geometry learned by OGP. RTA complements these losses by aligning distilled proxies with the relational topology of the transformed proxy identity space. Specifically, stacking private embeddings row-wise as $\hat{\mathbf{E}}$, we compute $\hat{\mathbf{E}}_{\mathrm{ort}}=\hat{\mathbf{E}}M_{\mathrm{ort}}^\top$ and apply K-Means per identity to extract $K$ prototypes, where $K$ matches the publication budget. This forms a normalized prototype matrix $\mathcal{P}\in\mathbb{R}^{NK\times D}$ as anchors of proxy identity geometry. Let $\mathbf{E}\in\mathbb{R}^{NK\times D}$ denote the normalized proxy embeddings in prototype order. We define the prototype topology and proxy-to-prototype affinity as
\begin{equation}
	\mathbf{A}^{*}=\mathcal{P}\mathcal{P}^{\top},
	\qquad
	\mathbf{A}^{\mathrm{syn}}=\mathbf{E}\mathcal{P}^{\top}.
\end{equation}
Since all embeddings are normalized, each entry is a cosine similarity. We align the target and proxy relation profiles by
\begin{equation}
	\label{eq:rta_structure}
	\begin{aligned}
		\mathcal{L}_{\mathrm{RTA}}&=\frac{1}{NK}\|(\mathbf{A}^{\mathrm{syn}}-\mathbf{A}^*)\odot\mathbf{I}\|_1 \\
		&+\frac{1}{NK(NK-1)}\|(\mathbf{A}^{\mathrm{syn}}-\mathbf{A}^*)\odot(\mathbf{1}-\mathbf{I})\|_1,
	\end{aligned}
\end{equation}
where $\mathbf{I}$ and $\mathbf{1}$ are the $NK\times NK$ identity and all-ones matrices, $\odot$ is the Hadamard product, and $\|\mathbf{X}\|_1=\sum_{i,j}|X_{ij}|$ is the entrywise $\ell_1$ norm. Separate normalization balances diagonal proxy-to-prototype alignment and off-diagonal intra- and inter-identity relations. Although $\mathbf{A}^{*}$ is orthogonally invariant, $\mathbf{A}^{\mathrm{syn}}$ uses the transformed prototypes $\mathcal{P}$, while the adapted backbone and proxy classifier anchor synthesis in the proxy space. The final distillation objective is
\begin{equation}
	\label{eq:total_loss}
	\mathcal{L}_{\mathrm{distill}} =	\mathcal{L}_{\mathrm{cls}}^{\mathrm{RTA}}+\lambda_{\mathrm{RTA}}\mathcal{L}_{\mathrm{RTA}}+\lambda_{\mathrm{bn}}\mathcal{L}_{\mathrm{bn}}.
\end{equation}
Through this objective, RTA transfers the proxy identity topology required for recognition learning to releasable images. This is a one-time publisher-side optimization, and after distillation, the released images require no private transformation, adapted model, or special inference pipeline.
\begin{table*}[t]
	\centering
	\resizebox{0.95\textwidth}{!}{
		\begin{tabular}{l|c|c|cccc|c}
			\toprule
			\textbf{Method} & \textbf{Type} & \textbf{Venue} & \textbf{Cross-Age} & \textbf{Surveillance} & \textbf{Large-Pose} & \textbf{VIS-NIR} & \textbf{Avg.} \\
			\midrule
			AdaFace \cite{Kim_2022_CVPR} & Base & CVPR'22 & 85.02 & 89.75 & \underline{88.97} & 95.00 & 89.69 \\
			Private Data Joint Training & Ref. & -- & 92.13 & 93.81 & 89.72 & 97.48 & 93.29 \\
			\midrule
			G$^2$Face~\cite{10644096} & \multirow{2}{*}{FA} & TIFS'24 & \underline{88.15} & 92.05 & 88.30 & 95.30 & \underline{90.95} \\
			NullFace~\cite{11557070} &  & FG'26 & 88.02 & \underline{92.36} & 87.32 & 95.35 & 90.76 \\
			\midrule
			DP-LoRA~\cite{Tsai_2025_ICCV} & \multirow{2}{*}{DP} & ICCV'25 & 85.62 & 91.41 & 83.98 & 92.35 & 88.34 \\
			Metric-DP~\cite{11146914} & & TPAMI'26 & 86.22 & 89.72 & 88.17 & 94.88 & 89.75 \\
			\midrule
			SRe$^2$L \cite{NEURIPS2023_e91fb65c} & \multirow{3}{*}{DD} & NeurIPS'23 & 85.02 & 91.78 & 88.12 & 94.48 & 89.85 \\
			NCFM \cite{Wang_2025_CVPR} &  & CVPR'25 & 84.25 & 91.64 & 88.20 & \underline{95.41} & 89.88 \\
			\rowcolor{gray!15} \textbf{Ours} & & -- & \textbf{88.42} & \textbf{93.69} & \textbf{89.09} & \textbf{95.43} & \textbf{91.66} \\
			\bottomrule
		\end{tabular}
	}
	\caption{Comparison of downstream FR adaptation performance~(\%). “Base” denotes the original pre-trained FR model, while “Ref.” denotes direct adaptation using private images. \textbf{Best} and \underline{second-best} results among publication methods and the baseline.}
	\label{tab:sota}
\end{table*}

\begin{figure*}[t]
	\centering
	\includegraphics[width=0.85\textwidth]{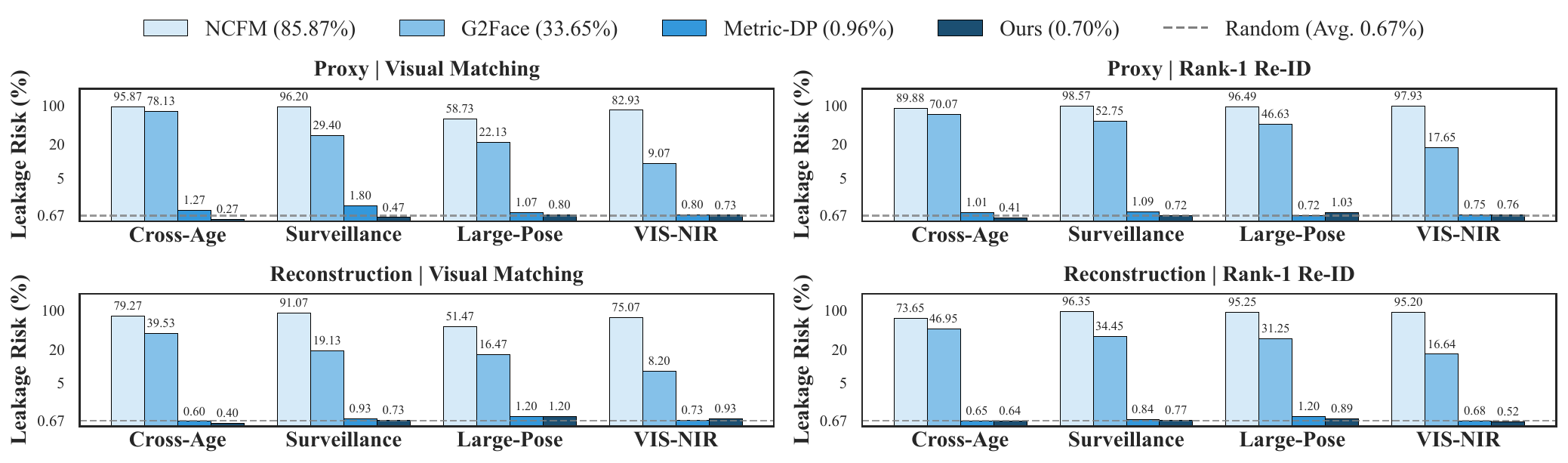}
	\caption{Source-identity linkability~(\%). Top: direct attacks on proxy datasets. Bottom: attacks on reconstructed images. Bars show Visual Matching (left) and Rank-1 Re-ID (right) success rates across target-domain scenarios. PFD averages 0.70\%, close to the 0.67\% random-guessing rate and the 0.96\% point estimate of the evaluated DP baseline.}
	\label{fig:bar}
\end{figure*}

\begin{table}
	\centering
	\setlength{\tabcolsep}{7pt}
	\renewcommand{\arraystretch}{1.1}
	\resizebox{0.95\columnwidth}{!}{
		\begin{tabular}{l|ccccc}
			\toprule
			\textbf{Method} & \textbf{LFW} & \textbf{CFP-FP} & \textbf{CPLFW} & \textbf{AgeDB} & \textbf{CALFW} \\
			\midrule
			G$^2$Face & 99.13 & 96.43 & 88.87 & 93.42 & 92.80 \\
			NullFace & 99.25 & \underline{96.87} & 89.53 & \underline{93.95} & \underline{93.06} \\
			DP-LoRA & 99.08 & 95.71 & 88.32 & 92.58 & 92.42 \\
			Metric-DP & \underline{99.35} & 96.86 & 89.60 & \underline{93.95} & 93.05 \\
			SRe$^2$L & 99.33 & 96.76 & \underline{89.67} & 93.60 & 92.82 \\
			NCFM & 99.32 & 96.50 & 89.27 & 93.42 & 92.98 \\
			\rowcolor{gray!15} \textbf{Ours} & \textbf{99.36} & \textbf{96.93} & \textbf{90.07} & \textbf{94.33} & \textbf{93.12} \\
			\bottomrule
		\end{tabular}
	}
	\caption{Comparison on standard FR benchmarks~(\%).}
	\label{tab:class}
\end{table}

\section{Experiments}
\label{sec:experiments}
\subsection{Datasets}
CASIA-WebFace~\cite{yi2014learning} is used as the public FR dataset for both pretraining and joint training with the published proxy dataset. We evaluate PFD on LFW~\cite{huang2008labeled}, CFP-FP~\cite{7477558}, CPLFW~\cite{zheng2018cross}, AgeDB~\cite{Moschoglou_2017_CVPR_Workshops}, CALFW~\cite{DBLP:journals/corr/abs-1708-08197} and our proposed Multi-Scenario Benchmark.

\subsection{Multi-Scenario Benchmark}
\label{sec:benchmark}
We construct a multi-scenario benchmark to evaluate downstream FR utility and source-identity linkability under private FR training dataset publication. Each scenario includes a private domain set for proxy generation and a disjoint test set for evaluation. For open-set evaluation, private identities are screened against CASIA-WebFace using feature-space similarity to reduce overlap. Each private set contains 150 identities with 50 images per identity, while each published set contains 150 arbitrary proxy classes with 10 images per class under a fixed budget. The source-to-proxy association is retained only for attack evaluation and is not released.

\subsubsection{Age Shift (Cross-Age).}
Derived from B3FD~\cite{B3FD}, this scenario evaluates cross-age FR adaptation under substantial age-related appearance variations. Evaluation is conducted on a disjoint test set following the standard LFW 1:1 verification protocol, using oldest-image anchors, genuine pairs from the same identity, and impostor pairs from different identities.

\subsubsection{Resolution Shift (Surveillance).}
Using the IJB-C dataset~\cite{8411217}, this scenario focuses on surveillance-domain FR adaptation with low-resolution and noisy imagery. A private surveillance subset is used for proxy distillation, while performance is evaluated on the remaining disjoint split under the official IJB-C protocol.

\subsubsection{Geometric Shift (Large-Pose).}
Built upon Multi-PIE~\cite{4813399}, this scenario emphasizes large-pose FR adaptation under extreme variations. The private split is biased toward challenging non-frontal views, while evaluation compares extreme-pose probes with identity-level cross-pose gallery templates.

\subsubsection{Modality Shift (VIS-NIR).}
Using LAMP-HQ~\cite{Yu2021}, this scenario studies cross-spectral FR adaptation between VIS and NIR domains. The private split maintains balanced VIS and NIR samples per identity, while evaluation is conducted on a disjoint VIS-NIR verification set following the same 1:1 verification protocol as LFW.

\begin{figure*}[!htbp]
	\centering
	\includegraphics[width=0.85\textwidth]{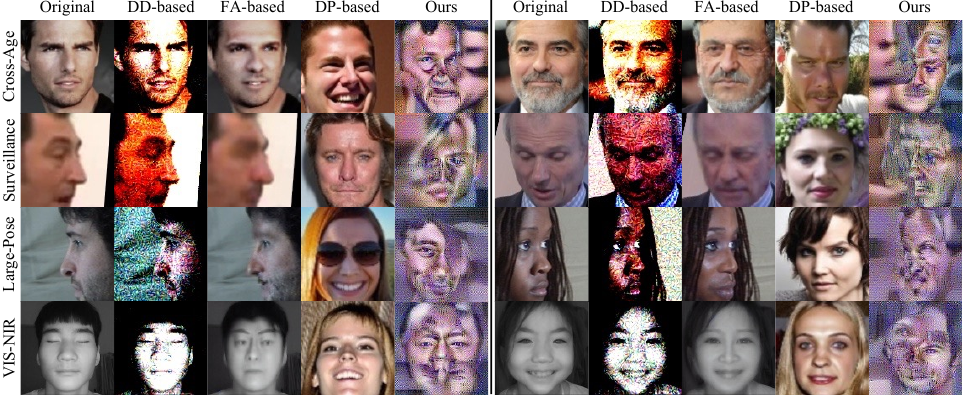}
	\caption{Visual comparison across four scenarios. Relative to the displayed DD-based~\cite{Wang_2025_CVPR}, FA-based~\cite{11557070}, and DP-based~\cite{11146914} methods, PFD produces more visually abstract proxies while retaining useful proxy classes for downstream FR adaptation.}
	\label{fig:dd_sample}
\end{figure*}

\begin{figure*}[!htbp]
	\centering
	\includegraphics[width=0.85\textwidth]{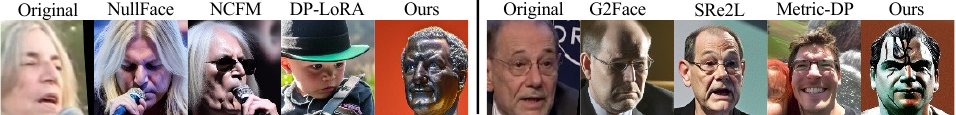}
	\caption{Representative post-release reconstruction stress test using off-the-shelf Arc2Face~\cite{10.1007/978-3-031-72913-3_14}. PFD reconstructions show weaker apparent source correspondence in these examples.}
	\label{fig:rec}
\end{figure*}

\subsection{Experiment Setting}
We use IR-50~\cite{Deng_2019_CVPR} as the backbone and AdaFace~\cite{Kim_2022_CVPR} as the FR classification loss. The public FR model is pre-trained on CASIA-WebFace. All face images are detected, aligned, and cropped to 112$\times$112 following~\cite{Kim_2022_CVPR}. FR model optimization is performed using SGD with cosine learning-rate decay and linear warmup, starting from 0.1. The proxy images are optimized from random noise using Adam with a learning rate of 0.25~\cite{NEURIPS2023_e91fb65c}. Training details are included in the Supplementary Materials.

\subsection{Evaluation Metrics}
\subsubsection{Downstream FR Utility.}
We evaluate downstream utility using scenario-specific FR protocols. For the Cross-Age and VIS-NIR scenarios, we report standard 1:1 verification accuracy. For the Surveillance scenario, we follow the official IJB-C protocol and report $\mathrm{TAR}@\mathrm{FAR}{=}1\text{e-}3$. For the Large-Pose scenario, we form genuine and impostor scores between extreme-pose probes and identity-level cross-pose gallery templates and report $\mathrm{TAR}@\mathrm{FAR}{=}1\text{e-}3$. For LFW-style datasets, we follow their official evaluation protocols.

\subsubsection{Source-Identity Linkability.}
To evaluate source-identity linkability under direct matching and reconstruction attacks, we follow prior privacy evaluation protocols~\cite{cherepanova2021lowkey} and report Visual Matching~(VisMatch) and Rank-1 Re-Identification~(Re-ID). Visual Matching associates each proxy or reconstruction with its nearest private source image, while Rank-1 Re-ID evaluates Top-1 retrieval of its associated source identity from the private gallery. Both report attack success when all 150 balanced source identities are provided as candidates, yielding a random-assignment rate of $1/150=0.67\%$. The source-to-proxy association is used only for scoring and remains hidden from the attacker.

\subsection{Utility and Source Linkability}
\subsubsection{Recognition Performance.} 
Tab.~\ref{tab:sota} compares PFD with representative FA-, DP-, and DD-based methods, as well as direct private-data joint training, under the same protocol. Existing methods weaken identity cues or alter facial appearance but lack an explicit mechanism for preserving transferable proxy identity geometry under domain shifts. PFD achieves the best performance across all scenarios, improving the average from 89.69\% to 91.66\%. It is also the only method that consistently surpasses the baseline, indicating stronger preservation of transferable identity structure. The advantage remains in the challenging Large-Pose setting, where PFD improves performance to 89.09\% despite severe pose variation. Results show that PFD preserves recognition-useful proxy geometry across domain shifts.

\begin{table*}[t]
	\centering
	\resizebox{0.98\linewidth}{!}{
		\begin{tabular}{lcc | ccc | ccc | ccc | ccc}
			\toprule
			\multirow{2}{*}{\textbf{Method}} & \multicolumn{2}{c|}{\textbf{Components}} & \multicolumn{3}{c|}{\textbf{Cross-Age}} & \multicolumn{3}{c|}{\textbf{Surveillance}} & \multicolumn{3}{c|}{\textbf{Large-Pose}} & \multicolumn{3}{c}{\textbf{VIS-NIR}} \\
			\cmidrule(lr){2-3} \cmidrule(lr){4-6} \cmidrule(l){7-9} \cmidrule(l){10-12} \cmidrule(l){13-15}
			& OGP & RTA & Perf. $\uparrow$ & VisMatch $\downarrow$ & Re-ID $\downarrow$ & Perf. $\uparrow$ & VisMatch $\downarrow$ & Re-ID $\downarrow$ & Perf. $\uparrow$ & VisMatch $\downarrow$ & Re-ID $\downarrow$ & Perf. $\uparrow$ & VisMatch $\downarrow$ & Re-ID $\downarrow$ \\
			\midrule
			Baseline & $\times$ & $\times$ & 85.02 & \underline{24.13} & \underline{28.61} & 91.78 & \underline{91.33} & \underline{93.80} & 88.12 & \underline{89.47} & \underline{93.68} & 94.48 & \underline{96.07} & \underline{96.80} \\
			+RTA  & $\times$ & \checkmark & \underline{86.20} & 46.00 & 64.85 & \underline{92.20} & 96.47 & 97.63 & \textbf{89.18} & 96.47 & 96.52 & \textbf{95.63} & 99.53 & 99.39 \\
			\rowcolor{gray!15} \textbf{Ours} & \checkmark & \checkmark & \textbf{88.42} & \textbf{0.27} & \textbf{0.41} &  \textbf{93.69} & \textbf{0.47} & \textbf{0.72} & \underline{89.09} & \textbf{0.80} & \textbf{1.03} & \underline{95.43} & \textbf{0.73} & \textbf{0.76} \\
			\bottomrule
		\end{tabular}
	}
	\caption{Ablation study of utility and source-identity linkability~(\%) across four target-domain adaptation scenarios. Performance is measured by Acc/TAR, while linkability is evaluated by VisMatch and Rank-1 Re-ID attack success rates.}
	\label{tab:ablation}
\end{table*}
To evaluate transfer beyond the scenario-specific test sets, Tab.~\ref{tab:class} reports results on standard FR benchmarks. We use the Large-Pose-adapted model for LFW, CFP-FP, and CPLFW, and the Cross-Age-adapted model for AgeDB and CALFW, following each benchmark's dominant variation. PFD obtains the highest reported score on all benchmarks and shows favorable transfer beyond the scenario-specific test sets.

\subsubsection{Source-Identity Linkability.}
Fig.~\ref{fig:bar} reports direct matching and Arc2Face reconstruction attacks using VisMatch and Rank-1 Re-ID. DD- and FA-based methods show substantially higher source linkage across scenarios. PFD achieves a mean attack success rate of 0.70\%, close to the 0.67\% chance rate and the 0.96\% DP baseline, while preserving stronger adaptation utility. Fig.~\ref{fig:dd_sample} and Fig.~\ref{fig:rec} visualize published proxies and their Arc2Face reconstructions. DD- and FA-based outputs retain more source-aligned appearance, whereas PFD produces abstract proxies with weaker apparent source correspondence after reconstruction. These observations agree with the quantitative linkage results in Fig.~\ref{fig:bar}, evaluated under the same matching protocols.

\subsubsection{Source-Attribute Recoverability.}
To assess demographic leakage, we use a fixed off-the-shelf attribute predictor~\cite{Karkkainen_2021_WACV} to compare race, gender, and age predictions between proxy images and their mapped sources. We report Macro-F1 and balanced accuracy to account for class imbalance. Since the source pseudo-labels span six observed age ranges, the age chance level is $1/6=16.67\%$. As shown in Tab.~\ref{tab:demographic}, average balanced accuracies are 23.53\%, 50.31\%, and 15.38\%, close to chance levels of 25.00\%, 50.00\%, and 16.67\%, indicating limited source-attribute recoverability under this predictor.

\begin{table}
	\centering
	\setlength{\tabcolsep}{7pt}
	\renewcommand{\arraystretch}{1.1}
	\resizebox{0.95\columnwidth}{!}{
		\begin{tabular}{l|ccc|ccc}
			\toprule
			\multirow{2}{*}{\textbf{Scenario}} & \multicolumn{3}{c}{\textbf{Macro-F1}} & \multicolumn{3}{c}{\textbf{Balanced Acc.}} \\
			\cmidrule(lr){2-4}\cmidrule(lr){5-7} & Race & Gender & Age & Race & Gender & Age \\
			\midrule
			Cross-Age & 22.01 & 54.60 & 16.12 & 21.22 & 54.73 & 18.73 \\
			Surveillance & 25.05 & 45.80 & 13.61 & 25.44 & 50.86 & 14.49 \\
			Large-Pose & 21.53 & 45.12 & 15.23 & 22.48 & 50.46 & 15.71 \\
			VIS-NIR & 17.21 & 38.86 & 7.98 & 24.98 & 45.20 & 12.57 \\
			\rowcolor{gray!15} \textbf{Avg.} & 21.45 & 46.10 & 13.24 & 23.53 & 50.31 & 15.38 \\
			\bottomrule
		\end{tabular}
	}
	\caption{Source-demographic attribute recoverability from released PFD proxies~(\%).}
	\label{tab:demographic}
\end{table}
\begin{figure}[t]
	\centering
	\includegraphics[width=0.7\columnwidth]{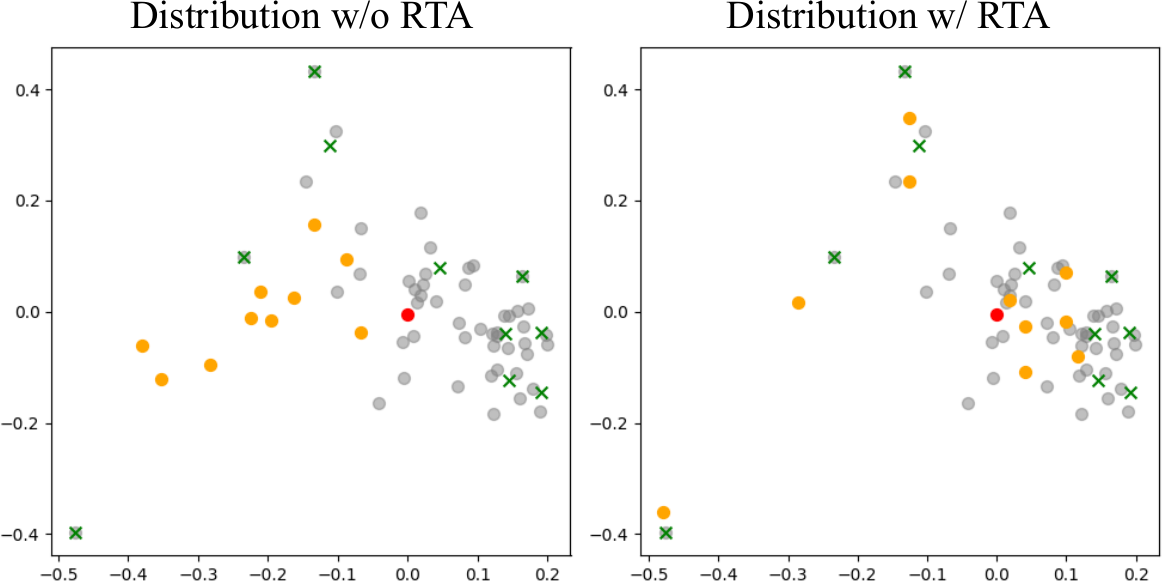}
	\caption{RTA aligns distilled proxies~(orange) with transformed prototypes~(green) to preserve the transferable relational geometry for downstream FR adaptation.}
	\label{fig:cluster}
\end{figure}
\subsection{Ablation Study}
We conduct incremental ablation studies to analyze the individual contributions of OGP and RTA to both target-domain adaptation utility and the reduction of source-identity linkability. Quantitative results are reported in Tab.~\ref{tab:ablation}, accompanied by qualitative visualizations for further insight.
\begin{figure}[t]
	\centering
	\includegraphics[width=0.75\columnwidth]{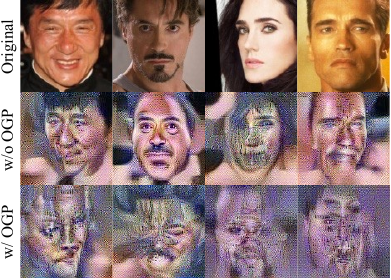}
	\caption{OGP reduces source-identity matching, yielding abstract proxies~(bottom) versus RTA-only~(middle).}
	\label{fig:ogp}
\end{figure}
\subsubsection{Relational Topology Alignment.} Adding RTA consistently improves target-domain adaptation performance over regular distillation. Fig.~\ref{fig:cluster} further shows that, without RTA, the distilled proxies tend to collapse into limited regions of the embedding space.  In contrast, RTA preserves both intra- and inter-identity relationships by explicitly aligning relational topology, leading to more discriminative and transferable proxy representations for downstream FR adaptation.

\subsubsection{Orthogonal Geometry Preservation.} OGP is designed to reduce source-identity linkability rather than directly improve adaptation utility. It weakens the correspondence between proxy and source identities while retaining recognition-useful identity geometry. Compared with the RTA-only variant, incorporating OGP reduces both VisMatch and Re-ID attack success from high levels to near chance across all scenarios while maintaining comparable adaptation performance. These results confirm the intended linkability-reduction role of OGP. Fig.~\ref{fig:ogp} provides the corresponding image-level comparison.

\section{Conclusion}
We study the identity paradox in private FR training dataset publication, where recognition-useful cues can also enable linkage to source identities. We propose Private Face Distillation to reduce source-aligned correspondence while preserving proxy identity geometry for recognition learning. Through Orthogonal Geometry Preservation and Relational Topology Alignment, PFD distills an identity-decoupled, geometry-preserving RGB proxy training set. Across domain shifts, the released proxies provide strong downstream utility and near-chance source linkage under the evaluated direct and reconstruction attacks.

\bibliography{aaai2027}

@inproceedings{NEURIPS2023_e91fb65c,
	author = {Yin, Zeyuan and Xing, Eric and Shen, Zhiqiang},
	booktitle = {Advances in Neural Information Processing Systems (NeurIPS)},
	pages = {73582-73603},
	title = {Squeeze, Recover and Relabel: Dataset Condensation at ImageNet Scale From A New Perspective},
	volume = {36},
	year = {2023}
}

@article{B3FD, 
	title={Picking out the bad apples: unsupervised biometric data filtering for refined age estimation},
	author={Be{\v{s}}eni{\'c}, Kre{\v{s}}imir and Ahlberg, J{\"o}rgen and Pand{\v{z}}i{\'c}, Igor S},
	journal={The visual computer},
	volume={39},
	number={1},
	pages={219-237},
	year={2023}
}

@INPROCEEDINGS{8411217,
	author={Maze, Brianna and Adams, Jocelyn and Duncan, James A. and Kalka, Nathan and Miller, Tim and Otto, Charles and Jain, Anil K. and Niggel, W. Tyler and Anderson, Janet and Cheney, Jordan and Grother, Patrick},
	booktitle={2018 International Conference on Biometrics (ICB)}, 
	title={IARPA Janus Benchmark - C: Face Dataset and Protocol}, 
	year={2018},
	volume={},
	number={},
	pages={158-165},
}

@InProceedings{Kim_2022_CVPR,
	author    = {Kim, Minchul and Jain, Anil K. and Liu, Xiaoming},
	title     = {AdaFace: Quality Adaptive Margin for Face Recognition},
	booktitle = {Proceedings of the IEEE/CVF Conference on Computer Vision and Pattern Recognition (CVPR)},
	year      = {2022},
	pages     = {18750-18759}
}

@InProceedings{Deng_2019_CVPR,
	author = {Deng, Jiankang and Guo, Jia and Xue, Niannan and Zafeiriou, Stefanos},
	title = {ArcFace: Additive Angular Margin Loss for Deep Face Recognition},
	booktitle = {Proceedings of the IEEE/CVF Conference on Computer Vision and Pattern Recognition (CVPR)},
	year = {2019}
}

@misc{yi2014learning,
	title={Learning Face Representation from Scratch}, 
	author={Dong Yi and Zhen Lei and Shengcai Liao and Stan Z. Li},
	year={2014},
	eprint={1411.7923},
	archivePrefix={arXiv},
	primaryClass={cs.CV},
	url={https://arxiv.org/abs/1411.7923}, 
}

@INPROCEEDINGS{4813399,
	author={Gross, Ralph and Matthews, Iain and Cohn, Jeffrey and Kanade, Takeo and Baker, Simon},
	booktitle={2008 8th IEEE International Conference on Automatic Face \& Gesture Recognition (FG)}, 
	title={Multi-PIE}, 
	year={2008},
	volume={},
	number={},
	pages={1-8},
}

@article{Yu2021, 
	author = {Yu, Aijing and Wu, Haoxue and Huang, Huaibo and Lei, Zhen and He, Ran}, 
	title = {LAMP-HQ: A Large-Scale Multi-pose High-Quality Database and Benchmark for NIR-VIS Face Recognition}, 
	journal = {International Journal of Computer Vision (IJCV)}, 
	volume={129},
	pages={1467–1483},
	year={2021},
}

@inproceedings{chan2025mgd3,
	author = {Chan Santiago, Jeffrey A. and Tirupattur, Praveen and Nayak, Gaurav Kumar and Liu, Gaowen and Shah, Mubarak},
	booktitle = {Proceedings of the 42nd International Conference on Machine Learning (ICML)},
	title = {{MGD}$^3$: Mode-Guided Dataset Distillation using Diffusion Models},
	year = {2025}
}

@misc{wang2018dataset,
	title={Dataset Distillation}, 
	author={Tongzhou Wang and Jun-Yan Zhu and Antonio Torralba and Alexei A. Efros},
	year={2020},
	eprint={1811.10959},
	archivePrefix={arXiv},
	primaryClass={cs.LG},
	url={https://arxiv.org/abs/1811.10959}, 
}

@article{10.1145/3708501,
	author = {Zhao, Ruoyu and Zhang, Yushu and Wang, Tao and Wen, Wenying and Xiang, Yong and Cao, Xiaochun},
	title = {Visual Content Privacy Protection: A Survey},
	year = {2025},
	publisher = {Association for Computing Machinery},
	address = {New York, NY, USA},
	volume = {57},
	number = {5},
	journal = {ACM Comput. Surv.},
	articleno = {122},
	numpages = {36},
}

@InProceedings{Cazenavette_2022_CVPR,
	author    = {Cazenavette, George and Wang, Tongzhou and Torralba, Antonio and Efros, Alexei A. and Zhu, Jun-Yan},
	title     = {Dataset Distillation by Matching Training Trajectories},
	booktitle = {Proceedings of the IEEE/CVF Conference on Computer Vision and Pattern Recognition (CVPR) Workshops},
	year      = {2022},
	pages     = {4750-4759}
}

@InProceedings{Wang_2025_CVPR,
	author    = {Wang, Shaobo and Yang, Yicun and Liu, Zhiyuan and Sun, Chenghao and Hu, Xuming and He, Conghui and Zhang, Linfeng},
	title     = {Dataset Distillation with Neural Characteristic Function: A Minmax Perspective},
	booktitle = {Proceedings of the IEEE/CVF Conference on Computer Vision and Pattern Recognition (CVPR)},
	year      = {2025},
	pages     = {25570-25580}
}

@ARTICLE{10914510,
	author={Zhao, Weisong and Zhu, Xiangyu and Shi, Haichao and Zhang, Xiao-Yu and Zhao, Guoying and Lei, Zhen},
	journal={IEEE Transactions on Image Processing (TIP)}, 
	title={Global Cross-Entropy Loss for Deep Face Recognition}, 
	year={2025},
	volume={34},
	number={},
	pages={1672-1685}
}

@inproceedings{cherepanova2021lowkey,
	title={LowKey: Leveraging Adversarial Attacks to Protect Social Media Users from Facial Recognition},
	author={Valeriia Cherepanova and Micah Goldblum and Harrison Foley and Shiyuan Duan and John P Dickerson and Gavin Taylor and Tom Goldstein},
	booktitle={International Conference on Learning Representations (ICLR)},
	year={2021},
	url={https://openreview.net/forum?id=hJmtwocEqzc}
}

@InProceedings{Sun_2020_CVPR,
	author = {Sun, Yifan and Cheng, Changmao and Zhang, Yuhan and Zhang, Chi and Zheng, Liang and Wang, Zhongdao and Wei, Yichen},
	title = {Circle Loss: A Unified Perspective of Pair Similarity Optimization},
	booktitle = {Proceedings of the IEEE/CVF Conference on Computer Vision and Pattern Recognition (CVPR)},
	year = {2020}
}

@InProceedings{Boutros_2022_CVPR,
	author    = {Boutros, Fadi and Damer, Naser and Kirchbuchner, Florian and Kuijper, Arjan},
	title     = {ElasticFace: Elastic Margin Loss for Deep Face Recognition},
	booktitle = {Proceedings of the IEEE/CVF Conference on Computer Vision and Pattern Recognition (CVPR) Workshops},
	year      = {2022},
	pages     = {1578-1587}
}

@InProceedings{pmlr-v97-lezcano-casado19a,
	title =    {Cheap Orthogonal Constraints in Neural Networks: A Simple Parametrization of the Orthogonal and Unitary Group},
	author =       {Lezcano-Casado, Mario and Mart\'{\i}nez-Rubio, David},
	booktitle =    {Proceedings of the 36th International Conference on Machine Learning (ICML)},
	pages =    {3794-3803},
	year =     {2019},
	volume =   {97},
	series =   {Proceedings of Machine Learning Research}
}

@inproceedings{daifracface,
	title={FracFace: Breaking The Visual Clues{\textemdash}Fractal-Based Privacy-Preserving Face Recognition},
	author={Wanying Dai and Beibei Li and Naipeng Dong and Guangdong Bai and Jin Song Dong},
	booktitle={The Thirty-ninth Annual Conference on Neural Information Processing Systems (NeurIPS)},
	year={2025},
	url={https://openreview.net/forum?id=JSSvYZKvL8}
}

@InProceedings{Mi_2023_ICCV,
	author    = {Mi, Yuxi and Huang, Yuge and Ji, Jiazhen and Zhao, Minyi and Wu, Jiaxiang and Xu, Xingkun and Ding, Shouhong and Zhou, Shuigeng},
	title     = {Privacy-Preserving Face Recognition Using Random Frequency Components},
	booktitle = {Proceedings of the IEEE/CVF International Conference on Computer Vision (ICCV)},
	year      = {2023},
	pages     = {19673-19684}
}

@InProceedings{Mi_2024_CVPR,
	author    = {Mi, Yuxi and Zhong, Zhizhou and Huang, Yuge and Ji, Jiazhen and Xu, Jianqing and Wang, Jun and Wang, Shaoming and Ding, Shouhong and Zhou, Shuigeng},
	title     = {Privacy-Preserving Face Recognition Using Trainable Feature Subtraction},
	booktitle = {Proceedings of the IEEE/CVF Conference on Computer Vision and Pattern Recognition (CVPR)},
	year      = {2024},
	pages     = {297-307}
}

@InProceedings{Wang_2023_CVPR,
	author    = {Wang, Zhibo and Wang, He and Jin, Shuaifan and Zhang, Wenwen and Hu, Jiahui and Wang, Yan and Sun, Peng and Yuan, Wei and Liu, Kaixin and Ren, Kui},
	title     = {Privacy-Preserving Adversarial Facial Features},
	booktitle = {Proceedings of the IEEE/CVF Conference on Computer Vision and Pattern Recognition (CVPR)},
	year      = {2023},
	pages     = {8212-8221}
}

@InProceedings{pmlr-v162-dong22c,
	title =    {Privacy for Free: How does Dataset Condensation Help Privacy?},
	author =       {Dong, Tian and Zhao, Bo and Lyu, Lingjuan},
	booktitle =    {Proceedings of the 39th International Conference on Machine Learning (ICML)},
	pages =    {5378--5396},
	year =     {2022},
	volume =   {162},
	series =   {Proceedings of Machine Learning Research},
	publisher =    {PMLR},
	url =      {https://proceedings.mlr.press/v162/dong22c.html},
}

@inproceedings{chen2025influenceguided,
	title={Influence-Guided Diffusion for Dataset Distillation},
	author={Mingyang Chen and Jiawei Du and Bo Huang and Yi Wang and Xiaobo Zhang and Wei Wang},
	booktitle={The Thirteenth International Conference on Learning Representations (ICLR)},
	year={2025},
	url={https://openreview.net/forum?id=0whx8MhysK}
}

@InProceedings{10.1007/978-3-031-72913-3_14,
	author={Papantoniou, Foivos Paraperas
	and Lattas, Alexandros
	and Moschoglou, Stylianos
	and Deng, Jiankang
	and Kainz, Bernhard
	and Zafeiriou, Stefanos},
	title={Arc2Face: A Foundation Model for ID-Consistent Human Faces},
	booktitle={Computer Vision -- ECCV 2024},
	year={2024},
	pages={241--261},
}

@ARTICLE{10644096,
	author={Yang, Haoxin and Xu, Xuemiao and Xu, Cheng and Zhang, Huaidong and Qin, Jing and Wang, Yi and Heng, Pheng-Ann and He, Shengfeng},
	journal={IEEE Transactions on Information Forensics and Security (TIFS)}, 
	title={G²Face: High-Fidelity Reversible Face Anonymization via Generative and Geometric Priors}, 
	year={2024},
	volume={19},
	number={},
	pages={8773-8785},
}

@INPROCEEDINGS{11557070,
	author={Kung, Han-Wei and Varanka, Tuomas and Sim, Terence and Sebe, Nicu},
	booktitle={2026 IEEE 20th International Conference on Automatic Face and Gesture Recognition (FG)}, 
	title={NullFace: Training-Free Localized Face Anonymization}, 
	year={2026},
	volume={},
	number={},
	pages={1-10},
}

@ARTICLE{11146914,
	author={Zhang, Yushu and Ji, Junhao and Wang, Tao and Zhao, Ruoyu and Wen, Wenying and Xiang, Yong},
	journal={IEEE Transactions on Pattern Analysis and Machine Intelligence (TPAMI)}, 
	title={Make Identity Indistinguishable: Utility-Preserving Face Dataset Publication With Provable Privacy Guarantees}, 
	year={2026},
	volume={48},
	number={1},
	pages={127-139},
}

@InProceedings{Tsai_2025_ICCV,
	author    = {Tsai, Yu-Lin and Li, Yizhe and Yu, Chia-Mu and Ren, Xuebin and Chen, Po-Yu and Chen, Zekai and Buet-Golfouse, Francois},
	title     = {Differentially Private Fine-Tuning of Diffusion Models},
	booktitle = {Proceedings of the IEEE/CVF International Conference on Computer Vision (ICCV)},
	year      = {2025},
	pages     = {4561-4571}
}

@inproceedings{huang2008labeled,
	title={Labeled faces in the wild: A database forstudying face recognition in unconstrained environments},
	author={Huang, Gary B and Mattar, Marwan and Berg, Tamara and Learned-Miller, Eric},
	booktitle={Workshop on faces in'Real-Life'Images: detection, alignment, and recognition},
	year={2008}
}

@INPROCEEDINGS{7477558,
	author={Sengupta, Soumyadip and Chen, Jun-Cheng and Castillo, Carlos and Patel, Vishal M. and Chellappa, Rama and Jacobs, David W.},
	booktitle={2016 IEEE Winter Conference on Applications of Computer Vision (WACV)}, 
	title={Frontal to profile face verification in the wild}, 
	year={2016},
	volume={},
	number={},
	pages={1-9},
}

@article{zheng2018cross,
	title={Cross-pose lfw: A database for studying cross-pose face recognition in unconstrained environments},
	author={Zheng, Tianyue and Deng, Weihong},
	journal={Beijing University of Posts and Telecommunications, Tech. Rep},
	volume={5},
	number={7},
	pages={5},
	year={2018}
}

@InProceedings{Moschoglou_2017_CVPR_Workshops,
	author = {Moschoglou, Stylianos and Papaioannou, Athanasios and Sagonas, Christos and Deng, Jiankang and Kotsia, Irene and Zafeiriou, Stefanos},
	title = {AgeDB: The First Manually Collected, In-The-Wild Age Database},
	booktitle = {Proceedings of the IEEE Conference on Computer Vision and Pattern Recognition (CVPR) Workshops},
	month = {July},
	year = {2017}
}

@article{DBLP:journals/corr/abs-1708-08197,
	publtype={informal},
	author={Tianyue Zheng and Weihong Deng and Jiani Hu},
	title={Cross-Age LFW: A Database for Studying Cross-Age Face Recognition in Unconstrained Environments},
	year={2017},
	cdate={1483228800000},
	journal={CoRR},
	volume={abs/1708.08197},
	url={http://arxiv.org/abs/1708.08197},
}

@inproceedings{meng2022improving,
	title={Improving Federated Learning Face Recognition via Privacy-Agnostic Clusters},
	author={Qiang Meng and Feng Zhou and Hainan Ren and Tianshu Feng and Guochao Liu and Yuanqing Lin},
	booktitle={International Conference on Learning Representations (ICLR)},
	year={2022},
}

@InProceedings{Chen_2022_CVPR,
	author    = {Chen, Jia-Wei and Yu, Chia-Mu and Kao, Ching-Chia and Pang, Tzai-Wei and Lu, Chun-Shien},
	title     = {DPGEN: Differentially Private Generative Energy-Guided Network for Natural Image Synthesis},
	booktitle = {Proceedings of the IEEE/CVF Conference on Computer Vision and Pattern Recognition (CVPR)},
	year      = {2022},
	pages     = {8387-8396}
}

@InProceedings{Barattin_2023_CVPR,
	author    = {Barattin, Simone and Tzelepis, Christos and Patras, Ioannis and Sebe, Nicu},
	title     = {Attribute-Preserving Face Dataset Anonymization via Latent Code Optimization},
	booktitle = {Proceedings of the IEEE/CVF Conference on Computer Vision and Pattern Recognition (CVPR)},
	year      = {2023},
	pages     = {8001-8010}
}

@InProceedings{Kuang_2024_CVPR,
	author    = {Kuang, Zhenzhong and Yang, Xiaochen and Shen, Yingjie and Hu, Chao and Yu, Jun},
	title     = {Facial Identity Anonymization via Intrinsic and Extrinsic Attention Distraction},
	booktitle = {Proceedings of the IEEE/CVF Conference on Computer Vision and Pattern Recognition (CVPR)},
	year      = {2024},
	pages     = {12406-12415}
}

@ARTICLE{10499238,
	author={Yuan, Lin and Chen, Wu and Pu, Xiao and Zhang, Yan and Li, Hongbo and Zhang, Yushu and Gao, Xinbo and Ebrahimi, Touradj},
	journal={IEEE Transactions on Information Forensics and Security (TIFS)}, 
	title={PRO-Face C: Privacy-Preserving Recognition of Obfuscated Face via Feature Compensation}, 
	year={2024},
	volume={19},
	number={},
	pages={4930-4944},
}

@inproceedings{10.1145/3503161.3548202,
	author = {Yuan, Lin and Liu, Linguo and Pu, Xiao and Li, Zhao and Li, Hongbo and Gao, Xinbo},
	title = {PRO-Face: A Generic Framework for Privacy-preserving Recognizable Obfuscation of Face Images},
	year = {2022},
	address = {New York, NY, USA},
	booktitle = {Proceedings of the 30th ACM International Conference on Multimedia (ACMMM)},
	pages = {1661–1669},
	numpages = {9},
	series = {MM '22}
}

@ARTICLE{9481149,
	author={Meden, Blaž and Rot, Peter and Terhörst, Philipp and Damer, Naser and Kuijper, Arjan and Scheirer, Walter J. and Ross, Arun and Peer, Peter and Štruc, Vitomir},
	journal={IEEE Transactions on Information Forensics and Security (TIFS)}, 
	title={Privacy–Enhancing Face Biometrics: A Comprehensive Survey}, 
	year={2021},
	volume={16},
	number={},
	pages={4147-4183},
}

@InProceedings{Zhu_2020_CVPR,
	author = {Zhu, Yuqing and Yu, Xiang and Chandraker, Manmohan and Wang, Yu-Xiang},
	title = {Private-kNN: Practical Differential Privacy for Computer Vision},
	booktitle = {Proceedings of the IEEE/CVF Conference on Computer Vision and Pattern Recognition (CVPR)},
	year = {2020}
}

@inproceedings{10.1145/3474085.3475367,
	author = {Li, Jingzhi and Han, Lutong and Chen, Ruoyu and Zhang, Hua and Han, Bing and Wang, Lili and Cao, Xiaochun},
	title = {Identity-Preserving Face Anonymization via Adaptively Facial Attributes Obfuscation},
	year = {2021},
	publisher = {Association for Computing Machinery},
	address = {New York, NY, USA},
	booktitle = {Proceedings of the 29th ACM International Conference on Multimedia (ACMMM)},
	pages = {3891–3899},
	numpages = {9},
	location = {Virtual Event, China},
	series = {MM '21}
}

@inproceedings{NEURIPS2025_615ce9f0,
	author = {Li, Wenyuan and Li, Guang and Maeda, Keisuke and Ogawa, Takahiro and Haseyama, Miki},
	booktitle = {Advances in Neural Information Processing Systems (NeurIPS)},
	pages = {67427--67458},
	publisher = {Curran Associates, Inc.},
	title = {Hyperbolic Dataset Distillation},
	volume = {38},
	year = {2025}
}

@ARTICLE{11509640,
	author={Zhang, Jingxuan and Dai, Lei and Ye, Fei and Chen, Zhihua and Li, Ping and Yang, Xiaokang and Sheng, Bin},
	journal={IEEE Transactions on Pattern Analysis and Machine Intelligence (TPAMI)}, 
	title={Dataset Distillation via a Noise-Unconstrained Generative Model}, 
	year={2026},
	volume={},
	number={},
	pages={1-18},
}

@inproceedings{NEURIPS2025_062cbe9b,
	author = {Zhang, Tianshuo and Gao, Li and Peng, Siran and Zhu, Xiangyu and Lei, Zhen},
	booktitle = {Advances in Neural Information Processing Systems (NeurIPS)},
	pages = {4103--4128},
	publisher = {Curran Associates, Inc.},
	title = {DevFD : Developmental Face Forgery Detection by Learning Shared and Orthogonal LoRA Subspaces},
	volume = {38},
	year = {2025}
}

@InProceedings{Peng_2026_CVPR,
	author    = {Peng, Siran and Zhang, Haoyuan and Gao, Li and Zhang, Tianshuo and Zhu, Xiangyu and Li, Bao and Zhao, Weisong and Lei, Zhen},
	title     = {DiffusionFF: A Diffusion-based Framework for Joint Face Forgery Detection and Fine-Grained Artifact Localization},
	booktitle = {Proceedings of the IEEE/CVF Conference on Computer Vision and Pattern Recognition (CVPR)},
	month     = {June},
	year      = {2026},
	pages     = {14095-14105}
}

@inproceedings{10.1145/3581783.3611711,
	author = {Zhao, Weisong and Zhu, Xiangyu and He, Zhixiang and Zhang, Xiao-Yu and Lei, Zhen},
	title = {Cross-Architecture Distillation for Face Recognition},
	year = {2023},
	publisher = {Association for Computing Machinery},
	address = {New York, NY, USA},
	booktitle = {Proceedings of the 31st ACM International Conference on Multimedia (ACMMM)},
	pages = {8076–8085},
	numpages = {10},
	location = {Ottawa ON, Canada},
	series = {MM '23}
}

@inproceedings{10.1145/3474085.3475464,
	author = {Kuang, Zhenzhong and Liu, Huigui and Yu, Jun and Tian, Aikui and Wang, Lei and Fan, Jianping and Babaguchi, Noboru},
	title = {Effective De-identification Generative Adversarial Network for Face Anonymization},
	year = {2021},
	publisher = {Association for Computing Machinery},
	booktitle = {Proceedings of the 29th ACM International Conference on Multimedia (ACMMM)},
	pages = {3182–3191},
	numpages = {10},
	location = {Virtual Event, China},
	series = {MM '21}
}

@InProceedings{Karkkainen_2021_WACV,
	author    = {Karkkainen, Kimmo and Joo, Jungseock},
	title     = {FairFace: Face Attribute Dataset for Balanced Race, Gender, and Age for Bias Measurement and Mitigation},
	booktitle = {Proceedings of the IEEE/CVF Winter Conference on Applications of Computer Vision (WACV)},
	month     = {January},
	year      = {2021},
	pages     = {1548-1558}
}


\end{document}